\documentclass[graybox]{svmult}

\usepackage{type1cm}        
\usepackage[bottom]{footmisc}
\usepackage{nicefrac}

\usepackage{natbib}         
							  
\usepackage[misc,geometry]{ifsym} 

\usepackage{etoolbox}
\newtoggle{extendedversion}
\toggletrue{extendedversion}

\usepackage{graphicx}
\usepackage{xcolor}
\usepackage{tikz}
\usepackage{pgfplots}
\usepgfplotslibrary{ternary, units}
\usepgfplotslibrary{
  colorbrewer,
}
\pgfplotsset{compat=1.10}
\usepgfplotslibrary{fillbetween}
\definecolor{winered}{rgb}{0.8,0,0}
\usetikzlibrary{patterns}
\usetikzlibrary{arrows}
\usetikzlibrary{arrows.meta}
\usepackage{amsmath,nccmath,bm}
\usepackage{mathrsfs}
\usepackage{amsfonts}
\usepackage{amssymb}
\usepackage{algorithm,algorithmicx}
\usepackage[noend]{algpseudocode}

\algrenewcommand\algorithmicrequire{\textbf{Precondition:}}
\algrenewcommand\algorithmicensure{\textbf{Postcondition:}}

\newcommand{\R}{\mathbb{R}}

\newcommand{\V}{\mathcal{V}}
\newcommand{\Q}{\mathcal{Q}}

\usepackage{bbm}
\newcommand{\EE}{\mathrm{E}}

\newcommand{\itr}{n}
\newcommand{\iter}{k}
\newcommand{\itrr}{m}
\newcommand{\iterr}{j}
\newcommand{\itrrr}{\ell}
\newcommand{\iterrr}{i}
\newcommand{\fromto}[2][1]{#1\kern-0.1em\colon\kern-0.2em #2}
\newcommand{\option}{u}

\newcommand{\dikP}{\Sigma}
\newcommand{\Ass}{\mathcal{A}}
\newcommand{\Echoice}[1][\Ass]{\smash{C^{\EE}_{\mathcal{P}(\Ass)}}}

\newcommand{\primal}{\text{\upshape{\textsc{Primal}}}}
\newcommand{\dual}{\text{\upshape{\textsc{Dual}}}}

\newcommand{\True}{\text{\upshape{{True}}}}
\newcommand{\False}{\text{\upshape{{False}}}}
\makeatletter
\newcommand{\labeltext}[3][]{%
    \@bsphack%
    \csname phantomsection\endcsname
    \def\tst{#1}%
    \def\labelmarkup{\emph}
    \def\refmarkup{}%
    \ifx\tst\empty\def\@currentlabel{\refmarkup{#2}}{\label{#3}}%
    \else\def\@currentlabel{\refmarkup{#1}}{\label{#3}}\fi%
    \@esphack%
    \labelmarkup{#2}
}
\makeatother

\usepackage{mathtools}

\newenvironment{proofof}[1]{\par\noindent{\bfseries\upshape Proof of #1\ }}{\qed\medskip}

\usepackage{enumitem}

\usepackage{newtxmath}       

\usepackage[capitalise]{cleveref}

\DeclarePairedDelimiter{\set}{\{}{\}}
\newcommand{\cset}[3][]{\set[#1]{#2\colon#3}}

\begin{document}

\title*{Decision-making with E-admissibility given a finite assessment of choices}
\author{Arne Decadt \and Alexander Erreygers \and Jasper De Bock \and Gert de Cooman}
\institute{Arne Decadt, Alexander Erreygers, Jasper De Bock and Gert de Cooman\at Foundations Lab, Ghent University, \email{arne.decadt@ugent.be}}
%
%
\maketitle


\abstract{
Given information about which options a decision-maker definitely rejects from given finite sets of options, we study the implications for decision-making with E-admissibility.
This means that from any finite set of options, we reject those options that no probability mass function compatible with the given information gives the highest expected utility.
We use the mathematical framework of choice functions to specify choices and rejections, and specify the available information in the form of conditions on such functions.
We characterise the most conservative extension of the given information to a choice function that makes choices based on E-admissibility, and provide an algorithm that computes this extension by solving linear feasibility problems.
}

\section{Introduction}\label{sec:introduction}
A decision-maker's uncertainty is typically modelled by a probability measure, and it is often argued that her rational decisions maximise expected utility with respect to this probability measure.
However, she may not always have sufficient knowledge to come up with a unique and completely specified probability measure.
It is then often assumed, as a work-around, that there is some set of probability measures that describes her uncertainty.
In this setting, \emph{E-admissibility} is among the more popular criteria for making choices, as indicated by \cite{troffaes2007decision}.

In this paper, we study and propose an algorithm for decision-making based on this criterion, starting from a finite uncertainty assessment.
As E-admissibility is popular, we are not the first to try and deal with this.
\citet{utkin2005powerful} and \citet{kikuti2005partially} have gone before us, but their assessments essentially only deal with pairwise comparison of options, while we can handle comparisons between sets of options.
\cite{decadt2020inference} have also studied more general assessments, but for other decision criteria than E-admissibility.
In order to achieve this generality, we will use choice functions as tools to model the decision-making process, because they lead to a very general framework, as argued elsewhere by, for instance, \cite{seidenfeld2010coherent,de2019interpreting,de2020archimedean}.

\iftoggle{extendedversion}{}{Proofs of our results can be found in an extended version%
.%
\footnote{https://arxiv.org/abs/2204.07428}
}

\section{Setting \& choice functions}\label{sec:choicef}
A choice function is a function that, for any given set of options, selects some subset of them.
The set~\(\V\) collects all \emph{options} and \(\Q\) is the set of all non-empty finite subsets of~\(\V\).
Formally, a \emph{choice function}~\(C\) is then a map from~\(\Q\) to itself such that \(C(A)\subseteq A\) for all~\(A \in\Q\).
If \(C(A)\) is a singleton consisting of a single option~\(u\), this means that \(u\) is chosen from~\(A\).
If \(C(A)\) has more than one element, however, we don't take this to mean that all the options in~\(C(A)\) are chosen, but rather that the options in~\(A\setminus C(A)\) are rejected and that the model does not contain sufficient information to warrant making a choice between the remaining options in~\(C(A)\).
Depending on the desired behaviour, various axioms can be imposed, leading to different types of choice functions; see for example \cite{seidenfeld2010coherent,de2019interpreting,de2020archimedean}.
In this contribution we consider choice functions under E-admissibility, as introduced in \cref{sec:Ead}.

We furthermore assume that we have an uncertain experiment with \(\itr\) possible outcomes, and we order the set of all outcomes~\(\mathcal{X}\) as \(\set{x_1,\ldots,x_{\itr}}\).
We interpret an option~\(u\) as a function that maps each outcome~\(x\) in~\(\mathcal{X}\) to the real-valued utility~\(u(x)\) that we get when the outcome of the uncertain experiment turns out to be~\(x\).
So we take the set of all options~\(\V\) to be the real vector space of all real-valued maps on~\(\mathcal{X}\).


\section{E-admissibility}\label{sec:Ead}
A decision-maker's uncertainty about an experiment is typically modelled by means of a probability mass function~\(p\colon \mathcal{X} \to [0,1]\), which represents the probability of each outcome in~\(\mathcal{X}\); we will use \(\dikP\) to denote the set of all such probability mass functions on~\(\mathcal{X}\).
The standard way---see for example \cite[Chapter~5]{savage1972foundations}---to choose between options~\(u\) proceeds by maximising expected utility with respect to~\(p\), where the expected utility of an option~\(u\in\V\) is given by~\(\EE_p(u)\coloneqq\sum_{x\in\mathcal{X}}u(x)p(x)\).

For every probability mass function~\(p\in\dikP\), the resulting choice function~\(C_p\) that maximises expected utility is defined by
\begin{equation}\label{eq:defCp}
C_{p}(A)
\coloneqq\cset{u\in A}{(\forall a\in A)\EE_p(u)\geq \EE_p(a)}
\text{ for all~\(A\in\Q\).}
\end{equation}

It is, however, not always possible to pin down exact probabilities for the outcomes \citep[Chapter~1]{walley1991statistical}.
Yet, the decision-maker might have some knowledge about these probabilities, for example in terms of bounds on the probabilities of some events.
Such knowledge gives rise to a set of probability mass functions~\(\mathcal{P}\subseteq \Sigma\), called a \emph{credal set} \citep[Section~1.6.2]{levi1978indeterminate}. 
In this context, there need no longer be a unique expected utility and so the decision-maker cannot simply maximise it.
Several other decision criteria can then be used instead; \citet{troffaes2007decision} gives an overview.
One criterion that is often favoured is \emph{E-admissibility}: choose those options that maximise expected utility with respect to at least one of the probability mass functions~\(p\) in~\(\mathcal{P}\) \citep{levi1978indeterminate}.\footnote{Levi's original definition considered credal sets that are convex, whereas we do not require this. In fact one of the strengths of our approach is that an assessment can lead to non-convex credal sets; see the example in \cref{ex:example} further on.}
If \(\mathcal{P}\) is non-empty, the corresponding choice function~\(C^{\EE}_{\mathcal{P}}\) is defined by
\begin{equation}\label{eq:unieOverCp}
  C^{\EE}_{\mathcal{P}} (A) \coloneqq \bigcup_{p\in\mathcal{P}} C_p(A)\text{ for all }A\in\Q.
\end{equation}
It will prove useful to extend this definition to the case that \(\mathcal{P}=\emptyset\).
\cref{eq:unieOverCp} then yields that \(C^{\EE}_{\emptyset}(A)=\emptyset\) for all~\(A\in\Q\), so \(C^{\EE}_{\emptyset}\) is no longer a choice function.
In either case, it follows immediately from \cref{eq:defCp,eq:unieOverCp} that
\begin{equation}\label{eq:oorsprdef}
C^{\EE}_{\mathcal{P}} (A)
=\cset{u\in A}{(\exists p\in\mathcal{P})(\forall a\in A) \EE_p(u)\geq \EE_p(a)}
\text{ for all~\(A\in\Q\).}
\end{equation}
The behaviour of choice functions under E-admissibility was first studied for horse lotteries by \cite{seidenfeld2010coherent}, characterised in a very general context by \cite{decooman2021banach}, and captured in axioms by \cite{de2020archimedean}.

\section{Assessments and extensions}
\label{sec:assessments}

We assume that there is some choice function~\(C\) that represents the decision-maker's preferences, but we may not fully know this function.
Our partial information about~\(C\) comes in the form of preferences regarding some---so not necessarily all---option sets.
More exactly, for some option sets~\(A\in\Q\), we know that the decision-maker rejects all options in~\(W\subseteq A\), meaning that \(C(A)\subseteq A\setminus W\); this can be also be stated as \(C(V\cup W)\subseteq V\), with \(V\coloneqq A\setminus W\).
We will represent such information by an \textit{assessment}: a set~\(\Ass\subseteq \Q^2\) of pairs~\((V,W)\) of disjoint option sets with the interpretation that, for all~\((V,W)\in\Ass\), the options in~\(W\) are definitely rejected from~\(V\cup W\).


Given such an assessment, it is natural to ask whether there is some choice function~\(C^{\EE}_{\mathcal{P}}\) under E-admissibility that agrees with it, in the sense that \(C^{\EE}_{\mathcal{P}}(V\cup W)\subseteq V\) for all~\((V,W)\in\Ass\).
Whenever this is the case, we call the assessment~\(\Ass\) \emph{consistent} with E-admissibility.
It follows from~\cref{eq:unieOverCp} that \(C^{\EE}_{\mathcal{P}}\) agrees with the assessment~\(\Ass\) if and only if
\[
\mathcal{P}
\subseteq\mathcal{P}(\Ass)
\coloneqq \cset[\big]{p\in\dikP}{(\forall(V,W)\in\Ass)C_p(V\cup W)\subseteq V}.
\]
Hence, \(\Ass\) is consistent if and only if \(\mathcal{P}(\Ass)\neq\emptyset\).
To check if \(\Ass\) is consistent, the following alternative characterisation will also be useful: for any \(A\in \Q\),
\begin{equation}\label{eq:testconsist}
\mathcal{P}(\Ass)\neq\emptyset
\Leftrightarrow \Echoice[\Ass](A)\neq \emptyset.
\end{equation}

If an assessment~\(\Ass\) is consistent and there is more than one choice function that agrees with it, the question remains which one we should use.
A careful decision-maker would only want to reject options if this is implied by the assessment.
So she wants a most conservative agreeing choice function under E-admissibility, one that rejects the fewest number of options.
Since larger credal sets lead to more conservative choice functions, this most conservative agreeing choice function under E-admissibility clearly exists, and is equal to~\(\Echoice[\Ass]\).
For this reason, we call~\(\Echoice[\Ass]\) the \emph{E-admissible extension} of the assessment~\(\Ass\).

So we conclude that checking the consistency of an assessment~\(\Ass\), as well as finding the E-admissible extension of a consistent assessment~\(\Ass\), amounts to evaluating~\(\Echoice[\Ass]\).
In the following sections we provide a method for doing this, which makes use of the following more practical expression for~\(\mathcal{P}(\Ass)\).


\begin{proposition}\label{prop:asss}
Consider an assessment~\(\Ass\).
Then \[\mathcal{P}(\Ass)=\cset[\big]{p\in\dikP}{(\forall (V,W)\in\Ass )(\forall w\in W)(\exists v\in V)\EE_p(v)>\EE_p(w)}.\]
\end{proposition}

\section{A characterisation of the E-admissible extension}\label{sec:characterisation}
Having defined the E-admissible extension~\(\Echoice[\Ass]\) of an assessment~\(\Ass\), it is only natural to wonder whether we can easily compute it.
We now turn to a method for doing so, albeit only for finite assessments.
As a first step, we derive a convenient characterisation of~\(\smash{\Echoice[\Ass]}\).
For any positive integer~\(\itrr\), it uses the notations~\(\smash{[\fromto{\itrr}]\coloneqq\set{1,\ldots,\itrr}}\) and \(d_{\fromto{\itrr}}\coloneqq(d_1,\ldots,d_{\itrr})\).

\begin{theorem}\label{th:E-adNat}
Consider an option set~\(A\), an option~\(\option\in A\) and a non-empty, finite assessment~\(\Ass\).
Enumerate the set~\(\cset{\cset{v-w}{v\in V}}{(V,W)\in\Ass, w\in W}\) as \(\set{D_1,\ldots,D_{\itrr}}\) and the set~\(\cset{\option-a}{a\in A\setminus\set{u}}\) as \(\set{u_1,\ldots,u_{\itrrr}}\).
Then \(\mathcal{P}(\Ass)=\bigcup_{d_{\fromto{\itrr}}\in\times_{\iterr=1}^\itrr D_{\iterr}} \mathcal{P}({d_{\fromto{\itrr}}}) \), where, for each~\(d_{\fromto{\itrr}}\in\times_{\iterr=1}^\itrr D_{\iterr}\), we let
\begin{equation}\label{eq:Pd1mdef}
\mathcal{P}({d_{\fromto{\itrr}}})\coloneqq \cset[\big]{p\in \Sigma}{(\forall \iterr\in [\fromto{\itrr}]) \EE_p(d_\iterr)> 0}.
\end{equation}
Furthermore, \(\option\in \Echoice(A)\) if and only if there is some~\(p\in\mathcal{P}(\Ass)\)---or equivalently some~\(d_{\fromto{\itrr}}\in\times_{\iterr=1}^\itrr D_{\iterr}\) and \(p\in \mathcal{P}({d_{\fromto{\itrr}}})\)---such that
\(
\EE_p(u_{\iterrr}) \geq 0\) for all \(\iterrr \in[\fromto{\itrrr}].
\)
\end{theorem}

\label{ex:example}
Let us illustrate the use of \cref{th:E-adNat} in determining, for a given option set, the resulting choices under the conservative E-admissible extension of a given assessment.
Let \(\mathcal{X}\coloneqq\set{1,2,3}\).
In order to allow for a graphical representation, we identify options and probability mass functions with vectors in~\(\R^3\), where for any~\(x\in\mathcal{X}\), the \(x\)-th component corresponds to the value of the option or probability mass function in~\(x\); so for example the option~\(w_1\coloneqq(1,-3,1)\) corresponds to the option that maps \(1\) to~\(1\), \(2\) to~\(-3\) and \(3\) to~\(1\).
We will choose from the option set~\(A\coloneqq\set{w_1,w_2,w_3}\), where we also let \(w_2\coloneqq(1,1,-2)\) and \(w_3\coloneqq(0,0,0)\).

For the assessment, we will consider \(v_1\coloneqq(-1,2,-2)\), \(v_2\coloneqq(-2,2,-1)\), \(v_3\coloneqq(0,3,-11)\), \(v_4\coloneqq(0,-7,-1)\), \(v_5\coloneqq(2,5,-9)\) and \(v_6\coloneqq(0,-2,-1)\).
Suppose that we are given the information that \(v_2\), \(v_3\) and \(v_4\) are rejected from~\(\set{v_1,v_2,v_3,v_4}\) and that \(v_1\) is rejected from~\(\set{v_1,v_5,v_6}\).
This corresponds to the assessment~\(\Ass=\set{{(\set{v_1},\set{v_2,v_3,v_4})},{(\set{v_5,v_6},\set{v_1})}}\).

\begin{figure}
\begin{center}\small
\begin{tikzpicture}[
    scale=.9,
    arrowhead/.style={->,>={Latex[winered,angle=60:7pt]}},
    blob/.style={ball color=winered,shape=circle,minimum size=5pt,inner sep=0pt},
    mycirc/.style={circle,fill=black,minimum size=0.001cm,scale=.5}
  ]
\begin{ternaryaxis}[
ternary limits relative=false,
xmin=0,
xmax=1,
ymin=0,
ymax=1,
zmin=0,
zmax=1,
xlabel=\(p(1)\),
ylabel=\(p(2)\),
zlabel=\(p(3)\),
label style={sloped},
minor tick num=1,
grid=both
]
\addplot3 [fill=blue!30,draw=none] table {
0.1 0.8 0.1
0.3 0.4 0.3
0.5 0.2 0.3
0.4 0.2 0.4
0.45 0.1 0.45
0.8 0.1 0.1
};

\addplot3 [fill=green!30,draw=none] table {
  0 0.6666 0.3333
  0.416666 0.25 0.3333
  0 0.25 0.75
};

\addplot3 [blue,very thick,dashed] coordinates {(0,0.7,0.3) (0.7,0,0.3)} node[near start,above,xshift=-0.5cm,sloped]{\(v_5-v_1\)};
\addplot3 [blue,very thick,dashed] coordinates {(0,0.2,0.8) (0.8,0.2,0)} node[near start,above,sloped]{\(v_6-v_1\)};

\addplot3 [green!60!black,very thick,dashed] coordinates {(0,2/3,1/3) (2/3,0,1/3)} node[near start,below,sloped]{\(w_3-w_2\)};
\addplot3 [green!60!black,very thick,dashed] coordinates {(0,1/4,3/4) (3/4,1/4,0)} node[near start,below,sloped]{\(w_3-w_1\)};
\addplot3 [green!60!black,very thick,dashed] coordinates {(1,0,0) (0,3/7,4/7)} node[near end,below,xshift=0.9cm, sloped]{\(w_1-w_2\)};

\addplot3 [winered,very thick,dashed] coordinates {(0.5,0,0.5) (0,1,0)} node[midway,above,sloped]{\(v_1-v_2\)};
\addplot3 [winered,very thick,dashed] coordinates {(0,0.9,0.1) (0.9,0,0.1)} node[midway,above,sloped]{\(v_1-v_3\)};
\addplot3 [winered,very thick,dashed] coordinates {(0.9,0.1,0) (0,0.1,0.9)} node[near end,above,sloped]{\(v_1-v_4\)};

\addplot3[black,thick] (0.6,0.15,0.25) node[mycirc,label=above:{\(p_1\)}]{};
\addplot3[black,thick] (0.6,0.2,0.2) node[mycirc,label=left:{\(p_2\)}]{};
\end{ternaryaxis}
\end{tikzpicture}
\end{center}
\caption{Ternary plot where the credal set~\(\mathcal{P}(\Ass)\) consists of those probability mass functions~\(p\colon\set{1,2,3}\to[0,1]\) that correspond to the blue region. A line labelled with an option~\(v\) means that \(\EE_p(v)=0\) for all~\(p\) on the line. The green region corresponds to the probability mass functions~\(p\) for which \(\EE_p(w_3-w_1)\geq 0\) and \(\EE_p(w_3-w_2)\geq 0\).}\label{fig:plotje}
\end{figure}
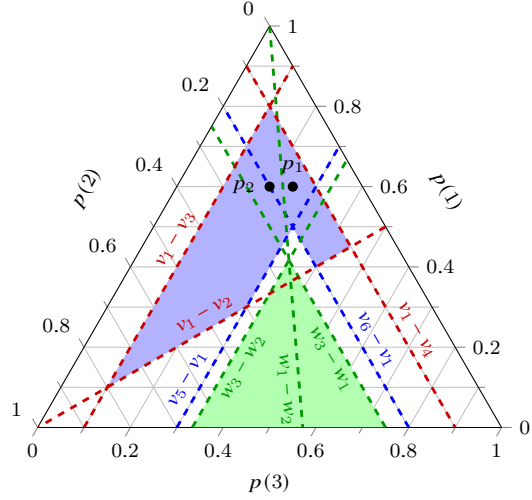

Now we will check for every option in~\(A\) whether it is in~\(\Echoice(A)\), by applying~\cref{th:E-adNat}.
For the sake of efficiency, we note that for all options, the assessment \(\mathcal{A}\) is the same, so they all have
\begin{multline*}
\set{D_1,\ldots,D_4}\\
\begin{aligned}
&=\set[\big]{\set{v_1-v_2},\set{v_1-v_3},\set{v_1-v_4},\set{v_5-v_1,v_6-v_1}}\\
&=\set[\big]{\set{(1,0,-1)},\set{(-1,-1,9)},\set{(-1,9,-1)},\set{(3,3,-7),(1,-4,1)}}.
\end{aligned}
\end{multline*}
In \cref{fig:plotje}, we have drawn the credal set~\(\mathcal{P}(\Ass)\) in blue in a ternary plot, using the characterisation in \cref{th:E-adNat}.


For~\(w_1\), the probability mass function~\(p_1\coloneqq(\nicefrac{12}{20},\nicefrac3{20},\nicefrac{5}{20})\) is consistent with the assessment, and we have \(\EE_{p_1}(u_1)=\nicefrac{3}{20}\geq0\) and \(\EE_{p_1}(u_2)=\nicefrac{2}{5}\geq0\), with $u_1\coloneqq w_1-w_2=(0,-4,3)$ and $u_2\coloneqq w_1-w_3=(1,-3,1)$.
Therefore, it follows from \cref{th:E-adNat} that \(w_1\) is not rejected from~\(A\) by~\(\Echoice[\Ass]\).
That \(w_2\) is not rejected either can be inferred similarly, for example using \(p_2\coloneqq(\nicefrac{3}{5},\nicefrac1{5},\nicefrac1{5})\).
For \(w_3\), we have \(u_1\coloneqq w_3-w_1=(-1,3,-1)\) and \(u_2\coloneqq w_3-w_2=(-1,-1,2)\).
The set of probability mass functions for which \(\EE_p(u_1)\geq0\) and \(\EE_p(u_2)\geq0\) corresponds to the green region in \cref{fig:plotje}, which has no overlap with the blue region.
Therefore, \(w_3\) is rejected from~\(A\) by~\(\Echoice[\Ass]\).
So we conclude that \(\Echoice[\Ass](A)=\{w_1,w_3\}\).

\section{An algorithmic approach}\label{sec:algorithm}
For larger problems, when the graphical approach in the example above is no longer feasible, we can translate \cref{th:E-adNat} into an algorithm.
A first way is to directly search, for each~\(d_{\fromto\itrr}\), for a probability mass function~\(p\) in~\(\mathcal{P}(d_{\fromto\itrr})\) such that \(u\in C_p(A)\).
To this end, we consider the function~\(\primal\colon \V^{\itrrr}\times \V^{\itrr} \to \set{\True,\False}\) that returns~\(\True\) if the following feasibility problem has a solution for input \((u_{\fromto{\itrrr}},d_{\fromto{\itrr}})\in\V^{\itrrr}\times \V^{\itrr}\) and \(\False\) otherwise, where \(\lambda_\iter\) can be seen as a scaled version of~\(p(x_\iter)\):
\begin{align*}
  \text{find}\quad        & \lambda_1,\dots,\lambda_{\itr}\in\R, \\
  \text{subject to\quad}  &  \textstyle\sum_{\iter=1}^{\itr}\lambda_{k} u_{\iterrr}(x_\iter) \geq 0  &&\text{for all } \iterrr\in[\fromto{\itrrr}],\\
                          &  \textstyle\sum_{\iter=1}^{\itr}\lambda_{k} d_{\iterr}(x_\iter) \geq 1  &&\text{for all } \iterr\in[\fromto{\itrr}],\\
                & \textstyle \lambda_{\iter}\geq 0    &&\text{for all } \iter\in[\fromto{\itr}].
\end{align*}

For another way to translate the condition in \cref{th:E-adNat} into an algorithm, we use duality.
That is, we consider the function
%
\(\dual\colon \V^{\itrrr}\times \V^{\itrr} \to \set{\True,\False}\) that returns \(\True\) if the following feasibility problem has a solution for input  \((u_{\fromto{\itrrr}},d_{\fromto{\itrr}})\in\V^{\itrrr}\times \V^{\itrr} \) and \(\False\) otherwise:
\begin{align*}
  \text{find}\quad        & \lambda_1,\dots,\lambda_{\itrrr+\itrr}\in\R, \\
  \text{subject to\quad}  &  \textstyle\sum_{\iterrr=1}^{\itrrr}\lambda_{\iterrr} u_\iterrr(x_\iter) +\sum_{\iterr=1}^{\itrr}\lambda_{\itrrr+\iterr} d_\iterr(x_\iter)\leq 0  &&\text{for all } \iter\in[\fromto{\itr}],\\
                & \textstyle  \sum_{\iterr=\itrrr+1}^{\itrrr+\itrr} \lambda_\iterr \geq 1,    &&\\
                & \textstyle \lambda_{\iterrr}\geq 0    &&\text{for all } \iterrr\in[\fromto{\itrrr+\itrr}].
\end{align*}
In practice, either of these feasibility problems can be solved by linear programming.\footnote{
It can for instance be considered as a linear programming problem, by adding the trivial objective function that is zero everywhere. Feeding this into a linear programming software package, the software will announce whether the problem is feasible.
For a deeper understanding of how software solves such feasibility problems, we refer to the explanation of initial feasible solutions in \citet[Section~5.6]{linprog2006}.}
Our next result relates these feasibility problems to the conditions in \cref{th:E-adNat}.

\begin{theorem}\label{th:2}
Consider option sequences~\(u_{\fromto{\itrrr}}\in\V^{\itrrr}\) and \(d_{\fromto{\itrr}}\in\V^{\itrr}\) and let \(\mathcal{P}(d_{\fromto\itrr})\) be as defined in \cref{eq:Pd1mdef}.
Then the following statements are equivalent:
\begin{enumerate}[label=\upshape(\roman*)]
  \item\label{itm:1} There is some~\(p\in\mathcal{P}({d_{\fromto{\itrr}}})\) such that \(\EE_p(u_{\iterrr}) \geq 0\) for all~\(\iterrr\in[\fromto{\itrrr}]\).
  \item\label{itm:2} \(\primal(u_{\fromto{\itrrr}},d_{\fromto{\itrr}})=\True\).
  \item\label{itm:3} \(\dual(u_{\fromto{\itrrr}},d_{\fromto{\itrr}})=\False\).
\end{enumerate}
\end{theorem}


\cref{th:E-adNat,th:2} guarantee that we can determine \(\Echoice(A)\) for any set~\(A\in\Q\) using~\cref{alg:IsAccepted}, by checking for each option~\(u\in A\) whether \(u\in \Echoice(A)\).
For any single~\(u\), this amounts to solving a linear feasibility program for each~\(d_{\fromto{\itrr}}\), using~\(\primal\) or~\(\dual\), as preferred.
Interestingly, consistency is equivalent to \(\Echoice(A)\neq \emptyset\), by \cref{eq:testconsist}.
In practice,  consistency can also be easily verified beforehand, by checking if \(0\in\Echoice[\Ass](\set{0})\), where `\(0\)' is the constant option that is \(0\) everywhere.
\vspace{-0.1cm}

\begin{algorithm}
\caption{Check for an option set~\(A\in\Q\) and a finite assessment~\(\Ass\) if an option~\(u\in A\) is in~\(\Echoice(A)\).\label{alg:IsAccepted}}
\begin{algorithmic}[1]
\Require{Let $\set{D_1,\ldots,D_{\itrr}}$ and \(\set{u_1,\ldots,u_{\itrrr}}\) be as in \cref{th:E-adNat}.}
\Statex
    \ForAll{$d_{\fromto{\itrr}} \in\smash{\times_{\iterr=1}^\itrr D_\iterr}$}
        \If{\(\primal(u_{\fromto{\itrrr}},d_{\fromto{\itrr}})\)} \Comment{Or \(\neg \dual(u_{\fromto{\itrrr}},d_{\fromto{\itrr}})\).}
            \State\Return{\(\True\)} \Comment{For one of the \(d_{\fromto{\itrr}}\) the condition is fulfilled.}
        \EndIf
    \EndFor
    \State\Return{\(\False\)} \Comment{When all elements of~\(\times_{j=1}^m D_j\) have been checked.}
\end{algorithmic}
\end{algorithm}
\vspace{-0.1cm}

If the assessment \(\Ass\) consists solely of pairs~\((V,W)\) where \(V\) is a singleton, then the corresponding set~\(\smash{\times_{\iterr=1}^\itrr D_\iterr}\) is a singleton, and the for-loop in \cref{alg:IsAccepted} vanishes.
Our algorithm can therefore be seen as repeatedly solving problems that have assessments of that form.

\vspace{-0.2cm}

\section{Conclusion}\label{sec:conclusion}
We have shown how to make choices using the E-admissibility criterion, starting from a finite assessment, using choice functions.
Our main conclusion is that calculating the most conservative E-admissible extension of this assessment reduces to checking linear feasibility multiple times.
Our setup is similar to the one previously studied by \citet{utkin2005powerful,kikuti2005partially}, the essential difference being that they have pairwise comparisons in the form of non-strict inequalities on expected utilities, whereas our assessments consider comparisons between sets of options, which leads to strict inequalities and allows for non-convex credal sets.
Future work could look into also implementing assessments that incorporate non-strict inequalities.
One way to do so would be through infinite assessments, so it might pay to look at which types of infinite assessments can still be handled finitely.

\iftoggle{extendedversion}{}{
\begin{acknowledgement}
The work of Jasper De Bock was supported by his BOF Starting Grant “Rational decision making under uncertainty: a new paradigm based on choice functions”, number 01N04819. We also thank the reviewers for their valuable feedback.
\end{acknowledgement}
}

	 
\bibliographystyle{spbasic}
\bibliography{Eadmilogica}

\iftoggle{extendedversion}{

\section*{Appendix}
\begin{proofof}{\cref{eq:testconsist}}
The reverse implication follows from \cref{eq:oorsprdef}.
To prove the direct implication, we assume that \(\mathcal{P}(\Ass)\neq \emptyset\) and fix any \(p\in \mathcal{P}(\Ass)\).
Assume \textit{ex absurdo} that \(C_p(A)\) is empty.
Then for every \(u\in A\) there is an \(a\in A\) such that \(\EE_p(a)> \EE_p(u)\).
But this would contradict \cite[I.3~Theorem~3]{birkhoff1940lattice} that says that there is a maximal element in \(A\) with respect to the partial order \(\succeq_p\coloneqq \cset{(x,y)\in\V^2}{\EE_p(x)>\EE_p(y) \vee x=y}\).
Hence, \(C^{\EE}_p(A)\neq \emptyset\).
Because \(p\in \mathcal{P}(\Ass)\), it follows from \cref{eq:unieOverCp} that \(C_{\mathcal{P}(\Ass)}(A)\neq \emptyset\), as required.
\end{proofof}

\begin{proofof}{\cref{prop:asss}}
By the definition of~\(\mathcal{P}(\Ass)\), we have to prove that for all~\(p\in\dikP\) and \((V,W)\in\Ass\) the following statements are equivalent:
\begin{equation}\label{eq:caral1}
 C_p(V\cup W)\subseteq V
\end{equation}
 and
 \begin{equation}\label{eq:caral2}
 (\forall w\in W)(\exists v\in V)\EE_p(v)>\EE_p(w).
 \end{equation}
 Take any~\(p\in\dikP\) and \((V,W)\in\Ass\).
First we prove that \cref{eq:caral1} implies \cref{eq:caral2}.
From \cref{eq:caral1} and the fact that \(V\) and \(W\) are disjoint, it follows that \(w\notin C_p(V\cup W)\) for all~\(w\in W\).
This means by definition that 
\begin{equation}\label{eq:oorsprk}
(\forall w\in W)(\exists a\in V\cup W)\EE_p(a)>\EE_p(w).
\end{equation}
We will now show that this implies \cref{eq:caral2}.
Take any option~\(w\in W\).
Let \(R\coloneqq\cset{r\in W}{r\succeq_p w}\), where we define the partial order \(\succeq_p\coloneqq \cset{(x,y)\in\V^2}{\EE_p(x)>\EE_p(y) \vee x=y}\).
Then by \cite[I.3~Theorem~3]{birkhoff1940lattice}, there is some maximal option~\(w^*\in R\) with respect to~\(\succeq_p\), since \(R\) is non-empty because it contains \(w\) and finite as it is a subset of~\(W\).
Since \(w^*\in R\subseteq W\), we know from \cref{eq:oorsprk} that there is some some~\(a^*\in V\cup W\) such that \(\EE_p(a^*)>\EE_p(w^*)\), and therefore also \(a^*\succeq_p w^*\) and \(a^*\neq w^*\).
Since \(w^*\) is maximal in~\(R\) with respect to~\(\succeq_p\), this implies that it is impossible that \(a^*\in R\).
It is also impossible that \(a^*\in W\setminus R\) because \(a^*\succeq_p w^*\succeq_p w\), where the second preference holds because \(w^*\in R\).
Hence, it must be that \(a^*\in V\).
Since \(W\) and \(V\) are disjoint, and \(w\in W\), this implies that \(a^*\neq w\). Since \(a^*\succeq_p w\), it follows that \(\EE_p(a^*)>\EE_p(w)\). So we have found some \(a^*\) in \(V\) such that \(\EE_p(a^*)>\EE_p(w)\).
As this holds for any option~\(w\in W\), we have proved~\cref{eq:caral2}.

Next we prove that \cref{eq:caral2} implies \cref{eq:caral1}.
Take any option~\(w \in W\).
Since \(V\subseteq V\cup W\), we have from \cref{eq:caral2} and the definition of~\(C_p\) that \(w\notin C_p(V\cup W)\).
Since this holds for any~\(w\in W\), it follows that \(C_p(V\cup W)\subseteq V\), and this is \cref{eq:caral1}.
\end{proofof}

\begin{proofof}{\cref{th:E-adNat}}
Let \(\mathcal{D}\coloneqq \smash{\times_{\iterr=1}^\itrr D_\iterr}\).
First we prove that 
\begin{equation}\label{eq:pAssss}
\mathcal{P}(\Ass)=\bigcup_{d_{\fromto{\itrr}}\in\mathcal{D}} \mathcal{P}({d_{\fromto{\itrr}}}). 
\end{equation}
By \cref{prop:asss} and \cref{eq:Pd1mdef}, this is equivalent to proving that for any \(p\in \dikP\) the following statements are equivalent
\begin{equation}\label{eq:asss}
(\forall (V,W)\in\Ass )(\forall w\in W)(\exists v\in V)\EE_p(v)>\EE_p(w)
\end{equation}
and 
\begin{equation}\label{eq:Pd1mdef2}
(\exists d_{\fromto{\itrr}}\in\mathcal{D}) (\forall \iterr\in [\fromto{\itrr}]) \EE_p(d_\iterr)> 0.
\end{equation}

First we prove that \cref{eq:asss} implies \cref{eq:Pd1mdef2}.
Take any \(\iterr\in [\fromto\itrr]\).
By definition of \(\{D_1,\dots,D_m\}\), there is a \((V,W)\in \Ass\) and a \(w\in W\) such that \(D_\iterr=\{v-w\colon v\in V\}\).
By \cref{eq:asss} there is a some~\(v^*\in V\) such that \(\EE_p(v^*)>\EE_p(w)\).
Let \(d_{\iterr}\coloneqq v^*-w\).
Then \(d_{\iterr}\in D_{\iterr}\) by definition.
From linearity of the expectation operator \(\EE_p\) and \(\EE_p(v^*)>\EE_p(w)\), it follows that \(\EE_p(d_{\iterr})=\EE_p(v^*-w)> 0\).
Since this holds for all \(\iterr\in [\fromto{\itrr}]\), \cref{eq:Pd1mdef2} holds.

Next we prove that \cref{eq:Pd1mdef2} implies \cref{eq:asss}.
Take some~\(d_{\fromto{\itrr}}\in\mathcal{D}\) that satisfies \cref{eq:Pd1mdef2}.
Take any \((V,W)\in \Ass\) and \(w\in W\).
Then by definition of \(\{D_1,\dots,D_m\}\), there is some~\(\iterr\in[\fromto{\itrr}]\) such that \(D_{\iterr}=\{v-w\colon v\in V\}\) and thus, since \(d_j\in D_j\), also some~\(v^*\in V\) such that \(d_{\iterr}=v^*-w\).
Then by \cref{eq:Pd1mdef2} we have \(\EE_p(v^*-w)=\EE_p(d_{\iterr})>0\) and from this and the linearity of the expectation operator \(\EE_p\), it follows that \(\EE_p(v^*)>\EE_p(w)\).
Since we can find such a \(v^*\) for any \((V,W)\in \Ass\) and \(w\in W\), we have proven \cref{eq:asss}.



For the second part of the statement, we rewrite \(\option\in \Echoice(A)\).
By \cref{eq:oorsprdef}, this is equivalent to the statement that there is some~\(p \in\mathcal{P}(\Ass)\) such that
\begin{equation}\label{eq:restat}
(\forall a\in A) \EE_p(u)\geq \EE_p(a).
\end{equation}
It therefore suffices to prove, for any \(p\in \dikP\), that \cref{eq:restat} is equivalent to
\begin{equation}\label{eq:stat2}
(\forall \iterrr\in[\fromto\itrrr]) \EE_p(u_\iterrr)\geq 0,
\end{equation}
which is what we now set out to do.

First we prove that \cref{eq:restat} implies \cref{eq:stat2}.
Take any \(\iterrr\in[\fromto\itrrr]\), then by definition of \(\{u_1,\dots,u_\ell\}\) there is some~\(a\in A\) such that \(u_{\iterrr}=u-a\).
By \cref{eq:restat}, \(\EE_p(u)\geq \EE_p(a)\) and by linearity of the expectation operator \(\EE_p\), we have \(\EE_p(u_{\iterrr})=\EE_p(u-a)\geq 0\).

Next we prove that \cref{eq:stat2} implies \cref{eq:restat}.
Take any \(a\in A\). If $a=u$, we trivially have that \(\EE_p(u)\geq \EE_p(a)\). If $a\neq u$, then by definition of \(\{u_1,\dots,u_\ell\}\) there is some~\(\iterrr\in [\fromto\itrrr]\) such that \(u_{\iterrr}=u-a\), or \(u=u_{\iterrr}+a\).
By \cref{eq:stat2}, \(\EE_p(u_\iterrr)\geq0\) and by linearity of expectation we have \(\EE_p(u)=\EE_p(u_{\iterrr})+\EE_p(a)\geq \EE_p(a)\).
\end{proofof}

\begin{proofof}{\cref{th:2}}
First we prove that \ref{itm:1} implies \ref{itm:2}.
By \ref{itm:1}, there is a~\(p\in \mathcal{P}(d_{\fromto\itrr})\) such that \(\EE_p(u_{\iterrr}) \geq 0\) for all~\(\iterrr\in[\fromto{\itrrr}]\).
By definition of \(\mathcal{P}(d_{\fromto\itrr})\), there is a~\(p\in\dikP\) such that \(\EE_p(d_{\iterr})>0\) for all~\(\iterr\in[\fromto{\itrr}]\) and \(\EE_p(u_{\iterrr}) \geq 0\) for all~\(\iterrr\in[\fromto{\itrrr}]\).
Then \(\eta\coloneqq \min_{\iterr\in[\fromto{\itrr}]} \EE_p(d_{\iterr})>0\).
For all~\(\iter \in[\fromto{\itr}]\), let \(\lambda_\iter\coloneqq \nicefrac{p(x_\iter)}{\eta}\geq 0\).
Then \(\sum_{\iter=1}^{\itr}\lambda_{\iter} u_{\iterrr}(x_{\iter})= \nicefrac{\EE_p(u_\iterrr)}{\eta} \geq 0 \text{ for all } \iterrr\in[\fromto{\itrrr}]\), because  \(\EE_p(u_\iterrr) \geq 0\) and \(\eta>0\), and \(\sum_{\iter=1}^{\itr}\lambda_{\iter} d_{\iterr}(x_\iter) = \nicefrac{\EE_p(d_{\iterr})}{\eta}\geq 1\) for all~\(\iterr\in[\fromto{\itrr}]\) because \(\EE_p(d_{\iterr})\geq \eta>0\).
In other words, the real numbers \(\lambda_1,\ldots,\lambda_{\itr}\) satisfy the primal linear feasibility problem, so \(\primal(u_{\fromto{\itrrr}},d_{\fromto{\itrr}})=\True\).

Second we prove that \ref{itm:2} implies \ref{itm:1}.
By \ref{itm:2}, there are real numbers \(\lambda_1,\ldots,\lambda_{\itr}\) such that \(\lambda_{\iter}\geq 0\) for all~\(\iter\in [\fromto{\itr}]\), \(\sum_{\iter=1}^{\itr}\lambda_{\iter} u_{\iterrr}(x_{\iter})\geq 0\) for all~\(\iterrr\in [\fromto{\itrrr}]\) and \(\sum_{\iter=1}^{\itr}\lambda_{\iter} d_{\iterr}(x_\iter)\geq 1\) for all~\(\iterr\in [\fromto{\itrr}]\).
Now let \(\eta\coloneqq \sum_{\iter=1}^{\itr} \lambda_{\iter}\).
Then \(\eta>0\).
To see why, assume \textit{ex absurdo} that \(\eta\leq 0\).
Then for all~\(\iter\in [\fromto{\itr}]\),  \(\lambda_{\iter}\geq 0\)  and \(\lambda_{\iter}=\eta -\sum_{\substack{\iter'\in[\fromto{\itr}]\setminus\{\iter\}}} \lambda_{\iter'}\leq 0\), and therefore \(\lambda_{\iter}= 0\), but this would imply that \(\sum_{\iter=1}^{\itr}\lambda_{\iter} d_{\iterr}(x_\iter)=0 \not\geq 1\), which is a contradiction. Hence, indeed, \(\eta>0\). Now define \(p(x_{\iter})\coloneqq \nicefrac{\lambda_{\iter}}{\eta}\) for all~\(\iter \in[\fromto{\itr}]\).
Then \(p\in\Sigma\), because \(\sum_{\iter=1}^{\itr}p(x_{\iter})=1\) and \(p(x_{\iter})\geq 0\), and \( \EE_p(d_{\iterr}) = \nicefrac{\sum_{\iter=1}^{\itr}\lambda_{\iter} d_{\iterr}(x_\iter)}{\eta}\geq \frac{1}\eta >0\) for all~\(\iterr\in[\fromto{\itrr}]\) because \(\sum_{\iter=1}^{\itr}\lambda_{\iter} d_{\iterr}(x_\iter)\geq 1\) and \(\eta>0\). This means that \(p\in \mathcal{P}(d_{\fromto\itrr})\).
Furthermore, we also have that \( \EE_p(u_\iterrr)=\nicefrac{\sum_{\iter=1}^{\itr}\lambda_{\iter} u_{\iterrr}(x_{\iter})}{\eta} \geq 0 \text{ for all } \iterrr\in[\fromto{\itrrr}]\), because  \(\sum_{\iter=1}^{\itr}\lambda_{\iter} u_{\iterrr}(x_{\iter}) \geq 0\) and \(\eta>0\).

Next we prove that \ref{itm:2} is equivalent to \ref{itm:3}.
We will use bold letters for vectors and matrices.
Note that \(\primal(u_{\fromto{\itrrr}},d_{\fromto{\itrr}})=\True\) if and only if there is some~\(\boldsymbol{\lambda}\in\R^{\itr}\) such that \(\boldsymbol{\lambda}\geq\mathbf{0}\) and \(\mathbf{A}\boldsymbol{\lambda}\leq \mathbf{b}\), with
\[
  \mathbf{A}=\begin{pmatrix}
  &-u_1(x_1) &\cdots &-u_1(x_\itr)\\
  &\vdots & & \vdots \\
  &-u_\itrrr(x_1) &\cdots &-u_\itrrr(x_\itr)\\
  &-d_1(x_1) &\cdots &-d_1(x_\itr)\\
  &\vdots & & \vdots \\
  &-d_\itrr(x_1) &\cdots &-d_\itrr(x_\itr)\\
  \end{pmatrix}\text{ and }
  \mathbf{b}=\begin{pmatrix}
  0\\
  \vdots\\
  0\\
  -1\\
  \vdots\\
  -1
  \end{pmatrix}.
\]
Farkas's Lemma \cite[Proposition~6.4.3(ii)]{linprog2006} tells us that this is equivalent to the condition that all~\(\mathbf{y}\in\R^{\itrr+\itrrr}\) that satisfy \(\mathbf{y}\geq \mathbf{0}\) and \(\mathbf{y}^{\mathrm{T}} \mathbf{A}\geq \mathbf{0}^\mathrm{T}\) also satisfy \(\mathbf{y}^\mathrm{T}\mathbf{b}\geq 0\).
By propositional logic, this is equivalent to the fact that there is no \(\mathbf{y}\in\R^{\itrr+\itrrr}\) that satisfies \(\mathbf{y}\geq \mathbf{0}\), \(\mathbf{y}^{\mathrm{T}} \mathbf{A}\geq \mathbf{0}^\mathrm{T}\) and \(\mathbf{y}^\mathrm{T}\mathbf{b}< 0\).
Since multiplying \(\mathbf{y}\) with a positive scalar has no effect on the veracity of these inequalities, this is in turn equivalent to the fact that there is no \(\mathbf{y}\in\R^{\itrr+\itrrr}\) such that \(\mathbf{y}\geq \mathbf{0}\), \(\mathbf{y}^{\mathrm{T}} \mathbf{A}\geq \mathbf{0}^\mathrm{T}\) and \(\mathbf{y}^\mathrm{T}\mathbf{b}\leq -1\), which holds if and only if \(\dual(u_{\fromto{\itrrr}},d_{\fromto{\itrr}})=\False\).

\end{proofof}

}{}
\end{document}